\documentclass[fleqn,10pt]{thescipub} 
\graphicspath{{Figures/}}
\usepackage{color}
\usepackage{fancyhdr}
\usepackage[caption=false,font=footnotesize]{subfig}
\usepackage{times}
\usepackage{bm}
\usepackage{hoang}
\usepackage[utf8]{inputenc}
\usepackage[thinc]{esdiff}
\usepackage{amsthm}
\usepackage{amsfonts}
\usepackage{amssymb,bbm}
\usepackage{algorithm}
\usepackage{algpseudocode}
\usepackage{thmtools}
\usepackage{thm-restate}
\usepackage{xcolor}
\usepackage{url}
\usepackage{chngpage}
\usepackage{enumitem}
\usepackage{booktabs}
\usepackage{pbox}
\usepackage{caption}
\usepackage{subcaption}
\usepackage{mathtools}
\usepackage{float}
\usepackage{natbib} 
\usepackage{soul}
\usepackage[english]{babel} 
\usepackage{hyperref}
\emergencystretch=\maxdimen
\usepackage{indentfirst}
\definecolor{color3}{RGB}{25,72,126} 

\usepackage{hyperref} 
\hypersetup{hidelinks,colorlinks,breaklinks=true,urlcolor=black,citecolor=black,linkcolor=black,bookmarksopen=false,pdftitle={Title},pdfauthor={Author}}

\JournalInfo{Journal of Computer Science} 

\PaperTitle{Sparse Partial Optimal Transport via Quadratic Regularization} 
\Authors{Khang Tran\textsuperscript{1}\thanks{These authors contributed equally to this work.}, Khoa Nguyen\textsuperscript{2}\footnotemark[1], Anh Nguyen\textsuperscript{3}, Thong Huynh\textsuperscript{4}, Son Pham\textsuperscript{5}, Sy-Hoang Nguyen-Dang\textsuperscript{4}, Manh Pham\textsuperscript{6, 7}, Bang Vo\textsuperscript{1}, Mai Ngoc Tran\textsuperscript{8, 7}, Dung Luong\textsuperscript{9}} 
\affiliation{\textsuperscript{1}\textit{Department of Computer Science, Ho Chi Minh University of Science, Ho Chi Minh City, Vietnam;}} 
\affiliation{\textsuperscript{2}\textit{School of Science, Aalto University, Espoo, Finland;}} 
\affiliation{\textsuperscript{3}\textit{Department of Science, Lycée Français Alexandre Yersin de Hanoi, Ha Noi, Vietnam;}} 
\affiliation{\textsuperscript{4}\textit{Department of Math, High School for the Gifted, Ho Chi Minh City, Vietnam;}} 
\affiliation{\textsuperscript{5}\textit{Department of Electrical and Computer Engineering, University of Massachusetts Amherst, Massachusetts, USA;}} 
\affiliation{\textsuperscript{6}\textit{Department of Computer Science, Georgia Institute of Technology, Atlanta, Georgia, USA;}} 
\affiliation{\textsuperscript{7}\textit{Acuitas Education, Ho Chi Minh City, Vietnam;}} 
\affiliation{\textsuperscript{8}\textit{Department of Computer Science, Binh Duong University, Ho Chi Minh City, Vietnam;}} 
\affiliation{\textsuperscript{9}\textit{Department of R\&D, VietDynamic, Ho Chi Minh City, Vietnam;}} 

\newcommand{\keywordname}{Keywords} 

\Abstract{Partial Optimal Transport (POT) has recently emerged as a central tool in various Machine Learning (ML) applications. It lifts the stringent assumption of the conventional Optimal Transport (OT) that input measures are of equal masses, which is often not guaranteed in real-world datasets, and thus offers greater flexibility by permitting transport between unbalanced input measures. Nevertheless, existing major solvers for POT commonly rely on entropic regularization for acceleration and thus return dense transport plans, hindering the adoption of POT in various applications that favor sparsity. In this paper,  as an alternative approach to the entropic POT formulation in the literature, we propose a novel formulation of POT with quadratic regularization, hence termed quadratic regularized POT (QPOT), which induces sparsity to the transport plan and consequently facilitates the adoption of POT in many applications with sparsity requirements. Extensive experiments on synthetic and CIFAR-10 datasets, as well as real-world applications such as color transfer and domain adaptations, consistently demonstrate the improved sparsity and favorable performance of our proposed QPOT formulation.}

\begin{document}
\captionsetup[figure]{labelfont={bf},name={Fig.},labelsep=period}
\captionsetup[table]{labelfont={bf},name={Table},labelsep=period}
\flushbottom 

\maketitle 


\section{Introduction}
Optimal Transport (OT) \citep{kantorovich1942translocation, villani2008stability} has long been a well-established mathematical framework for comparing probability distributions by finding the minimum-cost solution for transporting mass between them. However, the high computational cost of OT remains a significant drawback, limiting its practicality in large-scale problems. In recent years, the introduction of regularizers, particularly the entropic regularizer \citep{cuturi2013sinkhorn}, emerged as an answer to speeding up OT and improving its scalability, garnering significant attention as a powerful tool in modern Machine Learning (ML) applications. OT applications now span many fields of ML, such as color transfer \citep{piticolortransfer, poulicolortransfer}, domain adaptation \citep{redkodomainadaptaion}, and dictionary learning \citep{rolet2016fast}. 
A primary constraint of OT is its stringent assumption that both the masses of the source and the target distribution are equal, which does not hold in many real-world datasets.
This becomes the primary bottleneck for the application of OT to various ML problems \citep{rubner2000, Pele2009FastAR, gamfortfastot, nguyen2023unbalanced}, where such an assumption is not guaranteed.

\par
Partial Optimal Transport (POT) \citep{chapel2020partial} was introduced to address the challenges posed by OT and entropic regularized OT, allowing mass to be transported between unbalanced distributions. POT achieves this by relaxing the marginal constraints that OT strictly imposes \citep{Figalli2010, Caffarelli2010FreeBI}, providing greater flexibility \citep{chapel2020partial} and improved robustness to outliers \citep{Le2021Robustness}. However, as observed by \citep{nguyen2024partial}, the current implementation of POT in \citep{chapel2020partial} is effectively reformulations of OT with extended cost matrices, resulting in a significant increase in computation cost.  Additionally, \citep{nguyen2024partial} highlighted that the original Sinkhorn algorithm, when naively used to solve POT, is strictly infeasible as it violates some of the foundational constraints of the problem and proposes a revised feasible Sinkhorn algorithm for POT with \textit{worsen} complexity. 

\par
In addition, the entropic regularizer, along with the Sinkhorn algorithm \citep{cuturi2013sinkhorn, nguyen2024partial} or relevant solvers \citep{bregman_pot}, has recently become prevalent in the optimization of OT and POT problems thanks to its computational acceleration. However, the usage of entropic regularizers induces strictly dense transport plans \citep{blondel-smooth-and-sparse}, which is undesirable for several reasons. First, dense transport plans potentially impose significant memory complexity, making them inefficient when scaling to higher dimensions and larger datasets \citep{peyré2020computationaloptimaltransport}. Second, the lack of sparsity reduces the interpretability of the solution, obscuring the key connections between the source and target distributions \citep{solomon2015convolutional}. Third, such dense transport plans are often sensitive to noise, which undermines the reliability of the result \citep{genevay2019samplecomplexitysinkhorndivergences}. Finally, dense transport plans enforced by the solvers may nullify the usage of POT in many applications \citep{PITIE2007123, courty2016optimal, muzellec2016tsallis}, favoring sparse solutions.

\textbf{Contribution:} Given the worsening complexity of the Sinkhorn algorithm for POT  and the lack of sparsity due to the entropic regularizer in the existing solvers \citep{nguyen2024partial}, there is a strong need for an alternative regularization approach that can accelerate POT while inducing sparsity. 
This work proposes and benchmarks the quadratic regularizer as an effective solution, offering improved performance and encouraging sparsity for the POT solution.  To this end, we propose the novel quadratic regularized POT (QPOT) formulation, which augments the objective of the POT problem with quadratic regularization. 
We empirically evaluate QPOT across extensive settings of synthetic and CIFAR-10 datasets and real-world applications, such as color transfer and domain adaptation. 
Using entropic regularized POT (EPOT) as the baseline, our benchmarks demonstrate that QPOT achieves superior sparsity, proving its effectiveness in comparison to EPOT.

\section{Related Works}
\par
Applications of OT span a wide range of fields. In economics, \citep{10173668} has shown that using OT in resource allocation for Cloud–Edge Collaborative IoT significantly reduces the average energy consumption and delay. In ML, \citep{7586038} has applied OT for domain adaptation on toys, challenging real visual adaptation examples, and has shown that the method consistently outperforms state-of-the-art approaches. \citep{bousquet2017optimal}, in addition, has studied unsupervised generative modeling in terms of the OT problem and has shown a better understanding of the commonly observed blurriness of images generated by variational auto-encoders. In deep learning, the use of OT to generalize a deep neural network was studied in \citep{9546990}, while \citep{9173689} has applied OT in deep learning approaches for accelerated MRI and was able to reconstruct high-resolution MR images. OT is also used in image processing, as \citep{rabin2015non, blondel2018smooth,rabin2014adaptive} have applied OT in addition to other methods to gain great results in color transfer tasks. In biology, OT was applied in \citep{schiebinger2019optimal} to study developmental time courses to infer ancestor-descendant fates and model the regulatory programs that underlie them.

\par However, the classical optimal transport problem aims to find a transportation map that keeps the total mass between two probability distributions, requiring their mass to be the same. In certain cases, this condition can be hard to achieve, therefore leading to the formulation of Partial Optimal Transport (POT). The first known study of POT in ML applications was by \citep{chapel2020partial}. The work showed that this method is efficient in scenarios where point clouds come from different domains or have different features. Similarly, \citep{qin2022rigid} proposed a point cloud registration algorithm based on partial optimal transport and showed that the proposed method achieves state-of-the-art registration results when dealing with point clouds with significant amounts of outliers and missing points.
\par To further improve the OT method, recent studies have been about adding a regularizer term to improve certain aspects of the algorithm. The work of \citep{cuturi2013sinkhorn}, for example, has added an entropic regularizer term into the initial problem and shown that the algorithm runs at a speed that is several orders of magnitude faster than that of transportation solvers. Further studies on the efficiency and convergence of the entropic regularized OT can also be found in \citep{lin2022efficiency, carlier2017convergence}.
\par Even though the entropic regularizer has proven to increase the computing speed significantly, the resulting transportation plan is usually dense. To resolve this problem, many works have been conducted on studying the application of the quadratic regularizer to the OT problem. The work in \citep{blondel-smooth-and-sparse} shows that the incorporation of the quadratic norm and group lasso regularizations has led to an improvement in the sparsity of the transport plans. Additionally, \citep{lorenz2021quadratically} investigated the problem of optimal transport in the so-called Kantorovich form and derived two algorithms to solve the dual problem of the regularized problem. Their experiments have shown that both methods perform well, even for small regularization parameters.
Similarly, \citep{nutz2024quadratically} has proven the existence of the solution of the dual problem for a general square-integrable cost and that the optimal support is indeed sparse for small regularization parameters in a continuous setting with quadratic cost. On the other hand, in \citep{nguyen2024unbalancedoptimaltransportgradient}, the quadratic regularizer was applied simultaneously with the KL divergence to fill the lack of sparse UOT literature. Furthermore, motivated by an application to a sparse mixture of experts,  \citep{liu2022sparsity} studied OT with explicit cardinality constraints on the transportation plan and showed that this framework is formally equivalent to using squared $k$-support norm regularization in the primal.
\par In terms of the computational methods, the two main approaches for solving OT are gradient-based methods and the Sinkhorn algorithm. The Sinkhorn algorithm has been studied in \citep{cuturi2013sinkhorn} and has shown great computational speed. On the other hand, gradient methods have been studied in \citep{nesterov2021primal, an2022efficient} and have shown that the proposed method achieves faster convergence and better accuracy with the same parameter compared to the Sinkhorn algorithm. For POT, \citep{nguyen2024partial} has shown that Sinkhorn is not feasible and proposes a rounding algorithm to resolve the problem. In UOT, the gradient method has been studied in \citep{nguyen2024unbalancedoptimaltransportgradient} in addition to the primal-dual theory.

\section{Problem Formulation}
\subsection{Optimal Transport}
Define the source and target distribution for the transport as $\mathbf{r}, \mathbf{c} \in \mathbb{R}^n_+$ with mass $\|\mathbf{r}\|_1 \geq 0$ and $\|\mathbf{c}\|_1 \geq 0$. The Optimal Transport aims to move the mass of $\mathbf{r}$ to $\mathbf{c}$ and the other way around with minimal cost. \citep{kantorovich1942translocation} hence reformulate OT as a Linear Programming (LP) problem. 
\begin{equation} \label{OT}
    \mathbf{OT}(r, c) = \min \langle \mathbf{C}, \mathbf{X} \rangle \quad  \text{s.t.} \quad \mathbf{X} \in \cT(\mathbf{r}, \mathbf{c})
\end{equation}
where $\mathbf{C}$ is the cost matrix of moving between $\mathbf{r}$ and $\mathbf{C}$, and $\cT(\mathbf{r}, \mathbf{c})$ being the set of solution $\mathbf{X}$ that minimize the overall cost, defined as
\begin{equation} \label{constraintOT}
    \cT(\mathbf{r}, \mathbf{c}) \coloneqq \left\{ \mathbf{X} \in \mathbb{R}^{n\times n}_+ : \mathbf{X}\mathbf{1}_n = \mathbf{r}, \mathbf{X}^T\mathbf{1}_n = \mathbf{c}\right\}
\end{equation}. 
Revise the formulation of OT, the two equalities of $\cT(\mathbf{r}, \mathbf{c})$ enforces a strict constraint where the transported mass must be equal to the mass of both distributions, leading to the requirement where both masses must also be equal for the calculation of OT.

\subsection{Partial Optimal Transport}
The Partial Optimal Transport (POT), as defined in \citep{chapel2020partial} and \citep{Le2021Robustness}, relaxes such constraint from OT by introducing a total mass $0 \leq s \leq \min\{\|\mathbf{r}\|_1, \|\mathbf{c}\|_1\}$ that is allowed to be transported between the two distributions. The introduction of $s$ hence reformulate \ref{OT} as:
\begin{align}
\label{POT}
&\textbf{POT}(\textbf{r}, \textbf{c}, s)=\min \langle \mathbf{C}, \mathbf{X} \rangle
\end{align}
such that $\mathbf{X} \in \mathcal{U}(\mathbf{r}, \mathbf{c}, s)$ where 
\begin{align*}
    \mathcal{U}(\mathbf{r}, \mathbf{c}, s) \coloneqq 
    \{ 
    &\mathbf{X} \in \mathbb{R}^{n\times n}_+, \, \mathbf{X}\mathbf{1}_n \leq \mathbf{r}, \mathbf{X}^T\mathbf{1}_n \leq \mathbf{c}, \notag\\ &\mathbf{1}_n^T\mathbf{X}\mathbf{1_n} = s 
    \}.
\end{align*}
To elaborate, the addition of $s$ transforms the previous equalities in \ref{constraintOT} into inequalities in \ref{POT}, allowing the mass of each distribution to be transported partially and hence allowing the masses to differ. The loosening of the original constraint also allows POT to be more flexible and encourages sparsity in the transport plan, accelerating computations while maintaining the key connections between the two distributions. 

\subsection{Regularizers}
To be able to modify Linear Programming problems to focus on achieving certain tasks (e.g, lower runtime, higher accuracy, etc), a novel approach is to add a regularizer term to the objective function.

For OT problems, the entropic regularizer has been applied to both OT and POT  \citep{clason2021entropic,cuturi2013sinkhorn} and has given some optimistic results. The Entropic regularized OT (EOT) problem is OT with the addition of the regularizer Kullback–Leibler divergence  \citep{shlens2014notes} to stabilize the computation and make it solvable using faster algorithms such as the Sinkhorn-Knopp algorithm. Its formula can be written as: 
\begin{align} \label{EOT}
    &\mathbf{EOT}(\textbf{r}, \textbf{c}, \varepsilon) 
        = \min \left[ \langle \mathbf{C}, \mathbf{X} \rangle - \dfrac{\varepsilon}{2}H(\textbf{X})\right] \notag \\
    \quad \\&\text{s.t.} \quad \mathbf{X} \in \cT(\mathbf{r}, \mathbf{c}) \notag
\end{align}
where:
\begin{align*}
     H(\textbf{X})=\sum_{i,j}\textbf{X}_{i,j}\log\textbf{X}_{i,j}-\textbf{X}_{i,j}
\end{align*}

\par
Entropic regularized partial OT (EPOT) is the combination of POT and EOT, where our objective function would be the same as Eq.(\ref{EOT}) but with the additional condition that $\mathbf{X} \in \mathcal{U}(\mathbf{r}, \mathbf{c}, s)$ which have been denoted in (\ref{POT}). Formally, it can be written as:
\begin{align} \label{EPOT}
    & \mathbf{EPOT}(\textbf{r}, \textbf{c}, \varepsilon) = \min \left[ \langle \mathbf{C}, \mathbf{X} \rangle - \dfrac{\varepsilon}{2}H(\textbf{X}) \right] \notag\\
    & \quad \text{s.t.} \quad \mathbf{X} \in \mathcal{U}(\mathbf{r}, \mathbf{c},s)
\end{align}
\par
The greatest reason why the entropic regularizer is favored for OT problems is due to its great algorithmic speed \citep{cuturi2013sinkhorn}. However, the limitation of the entropic regularizer is that it would result in a dense transport plan, leading to difficulties in interpreting the general pattern, which may lead to suboptimal solutions in applications where sparsity is important. Additionally, when the regularizer coefficient $\varepsilon$ becomes very small, the Sinkhorn iterations used to solve the entropic OT problem may converge very slowly or face stability issues due to extremely large or small entries in the transport matrix.
\par
To resolve this problem, we propose the quadratic regularized POT (QPOT), which is POT with the additional usage of the quadratic regularizer (also known as the $\ell_2$ norm). More formally, the formula for QPOT can be written as:
\begin{align} \label{QPOT}
    &\mathbf{QPOT}(r, c, s, \varepsilon) = \min \left[ \langle \mathbf{C}, \mathbf{X} \rangle + \dfrac{\varepsilon}{2}||\mathbf{X}||_F^{2} \right] \notag\\
    &\text{s.t.} \quad \mathbf{X} \in \mathcal{U}(\mathbf{r}, \mathbf{c},s)
\end{align}
where $\varepsilon$ is the regularizer coefficient and $||\mathbf{T}||_F$ is the Frobenius norm of the matrix $\mathbf{T}$ for all matrix $\mathbf{T}$.
\par
The proposed method can perform very well due to the simplicity in computations of the novel $\ell_2$ norm as well as its ability to handle outliers. These properties of the $\ell_2$ norm have also been studied in \citep{zhu2015combining} and \citep{zhang2019non} and have given satisfactory results in terms of sparsity.

\section{Methods and Implementation}
We conducted our experiments on a system powered by an i7-12700K processor with 64GB of DDR5 RAM, comparing the performance of QPOT and EPOT in terms of sparsity through numerical experiments, color transfer, and domain adaptation tasks. Our implementation leveraged Python's \textit{cvxpy} library, which provides a natural mathematical interface for expressing convex optimization problems. Additionally, the library offers access to multiple open-source solvers, making it well-suited for our experimental setup. 
\par 
We selected CLARABEL and SCS as the solvers for QPOT and EPOT, respectively. The CLARABEL solver (introduced by \cite{clarabel}), based on a primal-dual interior-point method, aligns effectively with the smooth and convex optimization landscape introduced by the quadratic regularizer. CLARABEL is particularly efficient for quadratic objective functions as it does not require epigraphical reformulation of the objective function, leading to faster computation times. It operates with a tight convergence tolerance of approximately $10^{-8}$ and $10^{-6}$, resulting in highly accurate solutions.
\par
Conversely, the SCS solver (introduced by \cite{scs}) proved more suitable for EPOT. SCS employs Douglas-Rachford splitting combined with acceleration techniques, such as Anderson acceleration from \citep{aa2020}, which enhance its flexibility and efficiency when handling smooth and strongly convex regularizers like the entropic regularizer. The SCS solver achieves a convergence tolerance in the range of $10^{-3}$ and $10^{-4}$, providing a balance between computational speed and accuracy, making it a practical choice for large-scale entropic regularized problems.
\section{Experiments setup and Results}
The performance of QPOT is benchmarked through three experiments, using EPOT as the baseline. The first experiment involves a numerical analysis on the CIFAR-10 dataset and multiple toy distributions. We also conduct additional experiments on real-world machine learning applications, such as color transfer and domain adaptation. In all settings, we solve QPOT and EPOT using Python's \textit{cvxpy} package. Furthermore, our experiment setup aims to study the performance of both regularized POTs and observe how they behave under various parameters.
\par
Two noteworthy parameters that are studied throughout all experiments are $\lambda$ and $\epsilon$. $\lambda$ is the proportion of mass to be transported in the process. In other words, let $\lambda = 0.6$ and $\|\mathbf{r}\|_1 = \|\mathbf{c}\|_1 = 1$, then $\lambda$ is used to simulate the unbalance between the mass of source and target distributions, in this case, the mass to be transported is $s = \lambda*\min(\|\mathbf{r}\|_1, \|\mathbf{c}\|_1) = 0.6$. On the other hand, $\epsilon$ is the regularizers' strength, where the higher the value of $\epsilon$, the greater the effect of the regularizers, which drives the solver to focus more on optimizing the regularization term rather than the cost of the transportation.
\par
We measured the performance of both QPOT and EPOT based on their sparsities, defined as the number of entries in the transport plan below a certain threshold. For all four experiments, we set the threshold at $1e-10$. As discussed, sparsity is beneficial in the optimal transport (OT) problem because it improves computation time and promotes key connections between source and target distributions. Moreover, for better visualization, we color the heatmaps of all transport plans white for values smaller than the threshold $1e-10$ and black for all other values.
\par
Since $\lambda$ being too small would lead to the sparsity of both methods being too close to 1 and hard to differ, we have chosen $\lambda$ to be 0.5, 0.6, 0.7, 0.8, 0.9, 0.95, and 0.99 which is equivalent to 50\%, 60\%, 70\%, 80\%, 90\%, 95\% and 99\% of the allowed mass to be transported. Furthermore, for each value of $\lambda$, we run $\varepsilon$ from $10^{-i}$ to $10^{-j}$. Where $i,j$ will be chosen to numerically fit with each experiment.

\subsection{Numerical experiments}
This section presents the performance of QPOT and EPOT on two datasets. The first dataset is synthetic data sampled based on several of the classic distributions, such as mixed Gaussian, Poisson, Gamma, etc. On the other hand, in the second setup, the masses are drawn from the CIFAR10 dataset. 

\subsubsection{Sparsity experiment on toy distributions}
\textbf{Sparsity experiment on multiple pairs of distributions}
\par
\begin{figure}[!ht]
    \centering  \includegraphics[width=\linewidth]{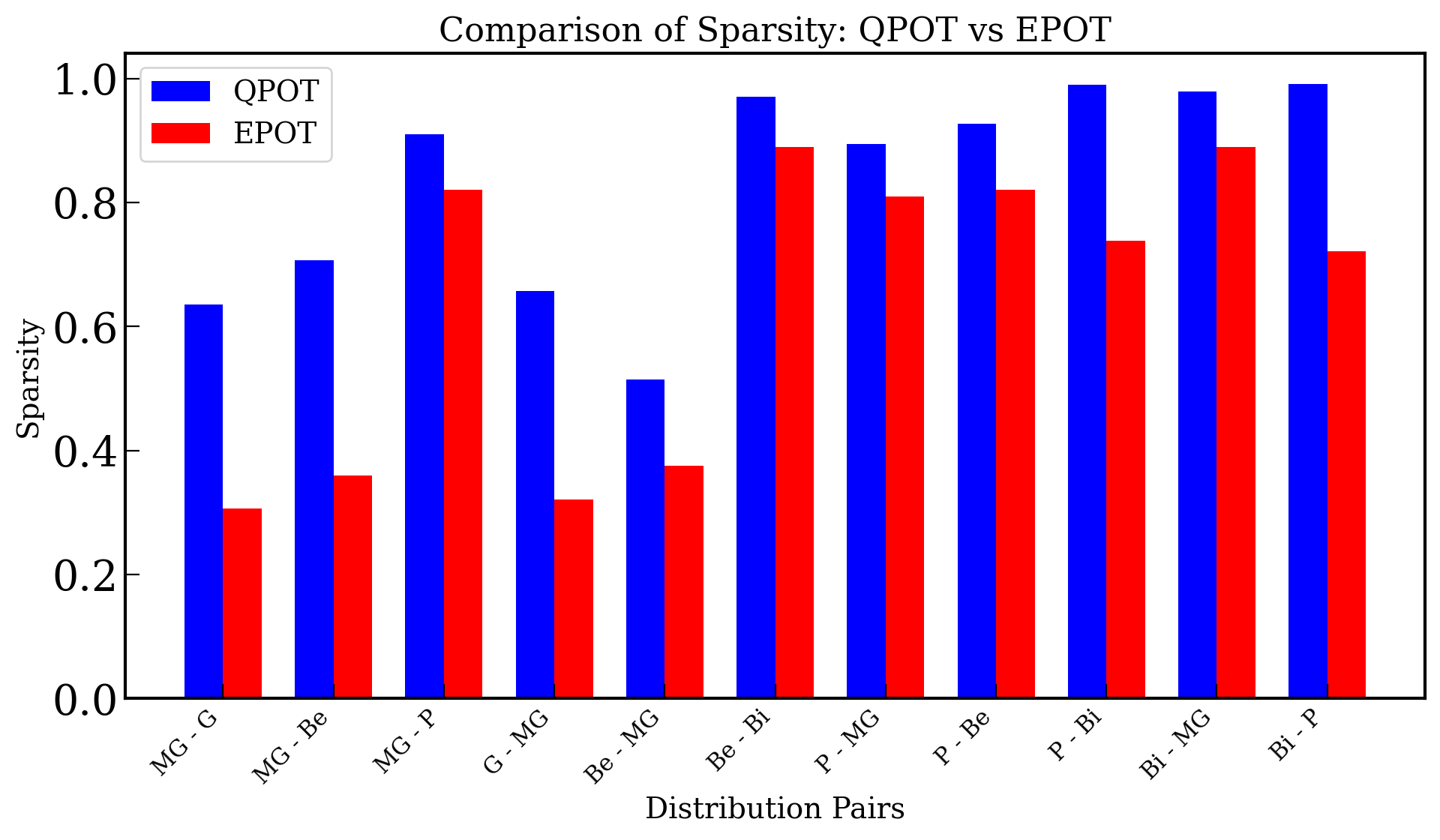}
    \caption{Comparing pairs of toy distributions}
    \label{fig:BestDisCompare}
\end{figure}

 The first setup compares the sparsities of QPOT and EPOT on multiple toy distributions in a one-to-one setting. We selected the pairs of distributions on several combinations of six different classic distributions and generated their mass by sampling them using Python's \textit{numpy} library. For each type of distribution, we generated $10^5$ samples, which are divided into 100 bins. Our sampled distributions include: the Gamma distribution (G): $\Gamma(\alpha=7, \beta=1) $, the Poisson distribution (P): $Pois(\lambda=5)$, the Binomial distribution $\beta(n=10, p=0.4$) (Bi):, the Beta distribution (Be): $Beta(\alpha=2,\beta=2)$, and the Mixed Gaussian distribution (MG): $\mathcal{N}_1(\mu_1=1, \sigma_1=2), \,\mathcal{N}_2(\mu_2=10, \sigma_2=1.5$). The results are plotted out in Figure \ref{fig:BestDisCompare}.

\par
It can be observed from Figure \ref{fig:BestDisCompare} that for all experimented pairs, QPOT has a more sparse transport plan compared to EPOT. Furthermore, for some pairs of toy distributions like MG-G or MG-Be, the sparsity of QPOT even doubles that of EPOT, which again highlights the dominance of our proposed method. \\

\textbf{Sparsity experiment on multiple mass}
\par The setting of $\lambda$ for this experiment is similar to what was mentioned at the beginning of this section. Additionally, we have chosen the regularizer coefficient $\varepsilon$ to be $1e-6$ and run our experiment on two toy distributions which are the Binomial Distribution $\beta(n=10, p=0.4$) and the Mixed Gaussian distribution $\mathcal{N}_1(\mu_1=1, \sigma_1=2),\, \mathcal{N}_2(\mu_2=10, \sigma_2=1.5)$. Figure \ref{fig:Lambda best} below shows the sparsity of both methods for each $\lambda$.
\begin{figure}[!ht]
    \centering
    \includegraphics[width=\linewidth]{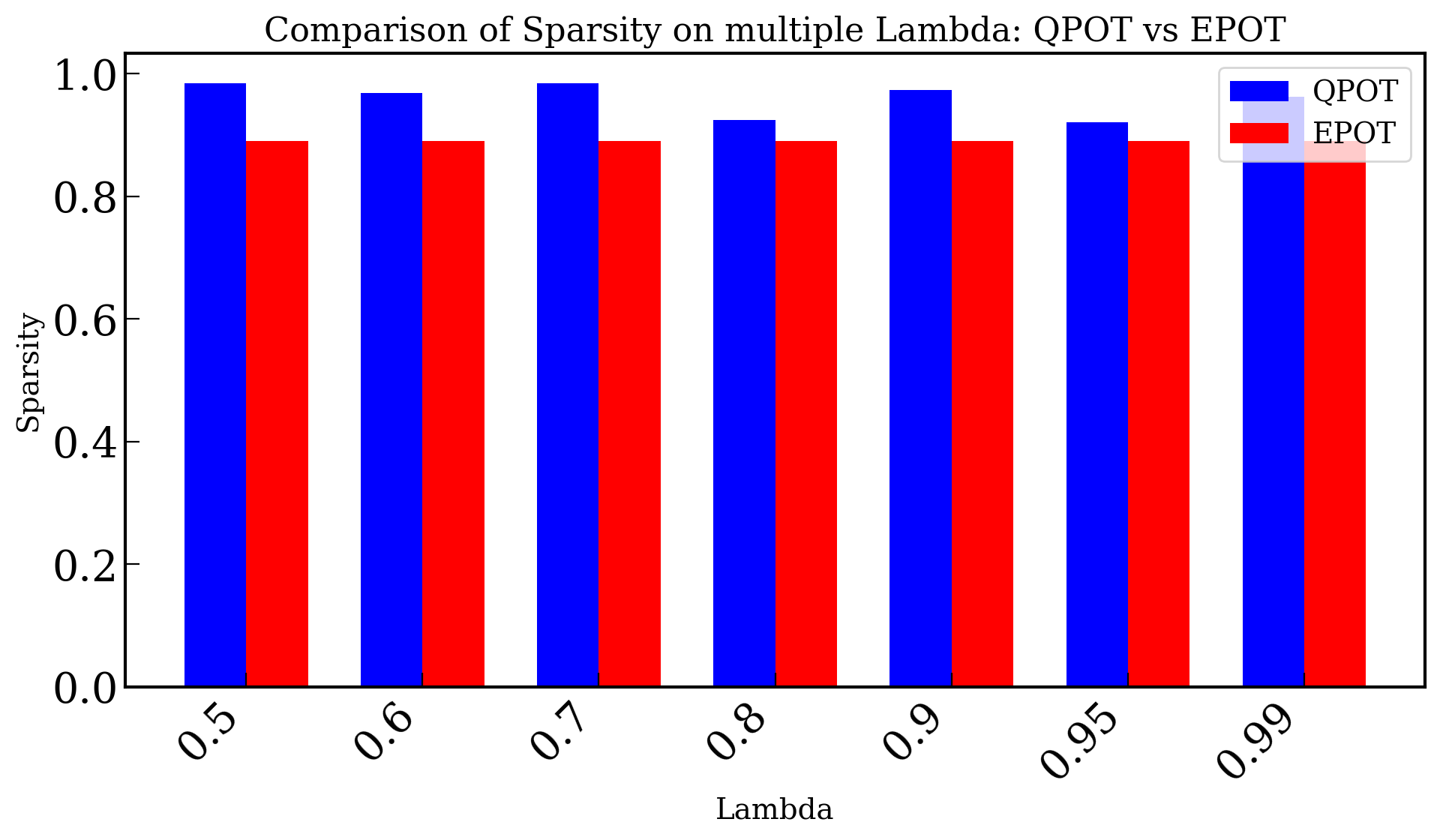}
    \caption{Sparsity of QPOT and QPOT for multiple transport masses}
    \label{fig:Lambda best}
\end{figure}
\par Similar to before, in all conducted experiments, QPOT  shows better performance in terms of sparsity compared to EPOT. Furthermore, overall values of $\lambda$, QPOT sparsity always stay higher than 0.9, while the same is not true for EPOT. This has confirmed the effectiveness of QPOT in finding a sparse solution even on different masses.

\begin{figure}[!ht]
\centering
\includegraphics[width=\linewidth]{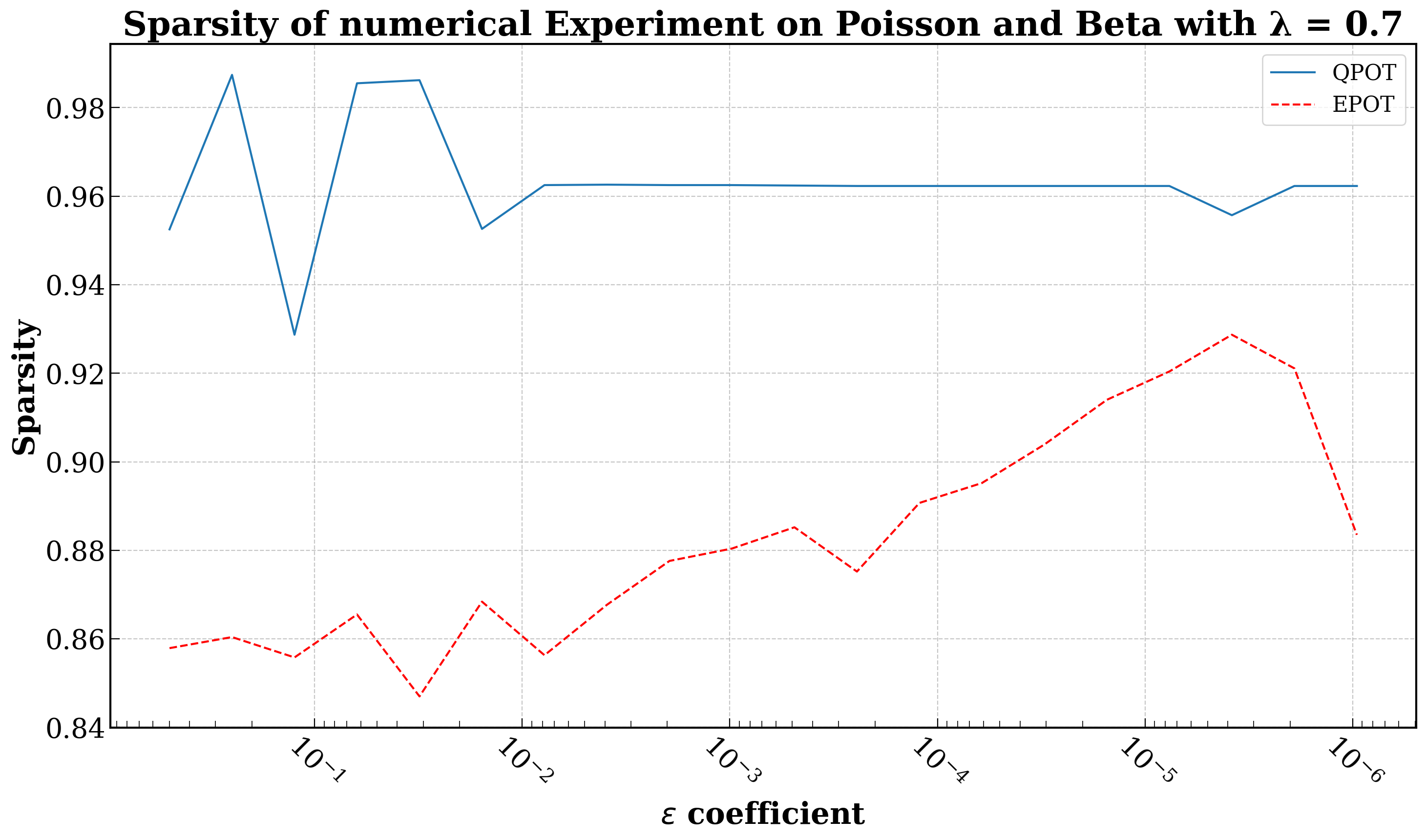}
\vspace{1em}
\includegraphics[width=\linewidth]{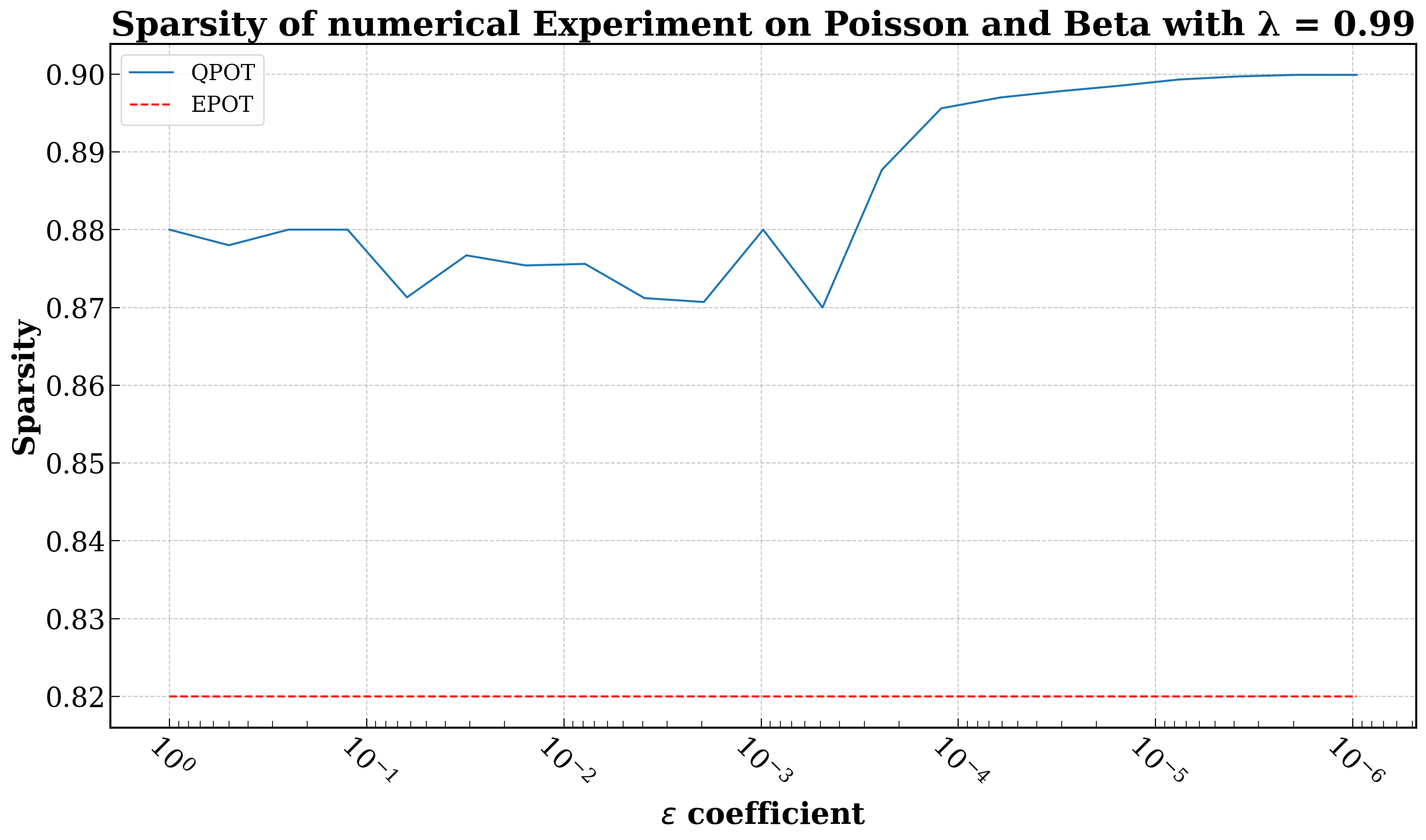}
\caption{Sparsity on Numerical Experiment on Poisson and Beta with: (a) $\lambda = 0.7$ (b) $\lambda = 0.99$}
\label{fig: toy_distribution_sparsity}
\end{figure}

\textbf{Sparsity experiment on multiple regularizer coefficients}\\
The next setup studies the effect of the regularizer strength on the performance of both EPOT and QPOT by varying the value $\epsilon$. As shown in Figure \ref{fig: toy_distribution_sparsity}, we benchmarked both POT methods on $\epsilon$ running from $10^{-0.5}$ to $10^{-6}$ with the pair of $Pois(\lambda=5)$ and $Beta(\alpha=2,\beta=2)$ are selected as the source and target distributions. We also studied this effect on two $\lambda$ values $0.99$ and $0.7$, as the first value simulates the setup of almost equal distribution, while it is studied that the sparsities of the two methods are the highest when $\lambda = 0.7$. Furthermore, similar to the previous experiment, for each type of distribution, we used $10^5$ samples to generate and divided them into 100 bins.

\par In both cases of $\lambda$, it can be seen that QPOT surpasses EPOT for all experimented $\epsilon$.  Additionally, it can also be observed that for most values of epsilon, the sparsity of EPOT is usually below 0.9, while the sparsity of QPOT mostly stays above 0.95. This, again, shows how dominant QPOT is when it comes to sparsity.
\subsection{Sparsity experiment on CIFAR10 dataset}
\par
Numerical experimentation on CIFAR-10 allows us to evaluate the optimal transport methods and on standard benchmark datasets where the masses are drawn from images. 
\par
The CIFAR-10 dataset is a well-established dataset that provides a diverse collection of 60,000 color images in RGB format, each sized $32\times 32$ pixels, divided into 10 object classes: airplane, automobile, bird, cat, deer, dog, frog, horse, ship, and truck. It includes 50,000 training images and 10,000 test images, making it a benchmark for many machine learning tasks that involve images. Despite its low resolution, CIFAR-10 remains challenging due to its diverse and complex visual content. The diversity in object classes, and consequently the differences in pixel value distribution images, makes CIFAR-10 an excellent candidate for this experiment. In this experiment, we used a pair of colored images from the CIFAR-10. We have decided to convert RGB images to grayscale images and to downscale the dimensions of the images from 32 x 32 to 10 x 10. The grayscale-reduced images are flattened into histograms with 100 bins. 
\par
We smooth and normalize both histograms by adding a small value of ${10}^{-6}$ to each bin. We calculate the cost matrix C using the squared Euclidean distance between pixel positions and normalize it so that the maximum value is $1$ ($\|C\|_{max} =1$). 
\par
\begin{figure}[!httbp]
    \centering
    \includegraphics[width=\linewidth]{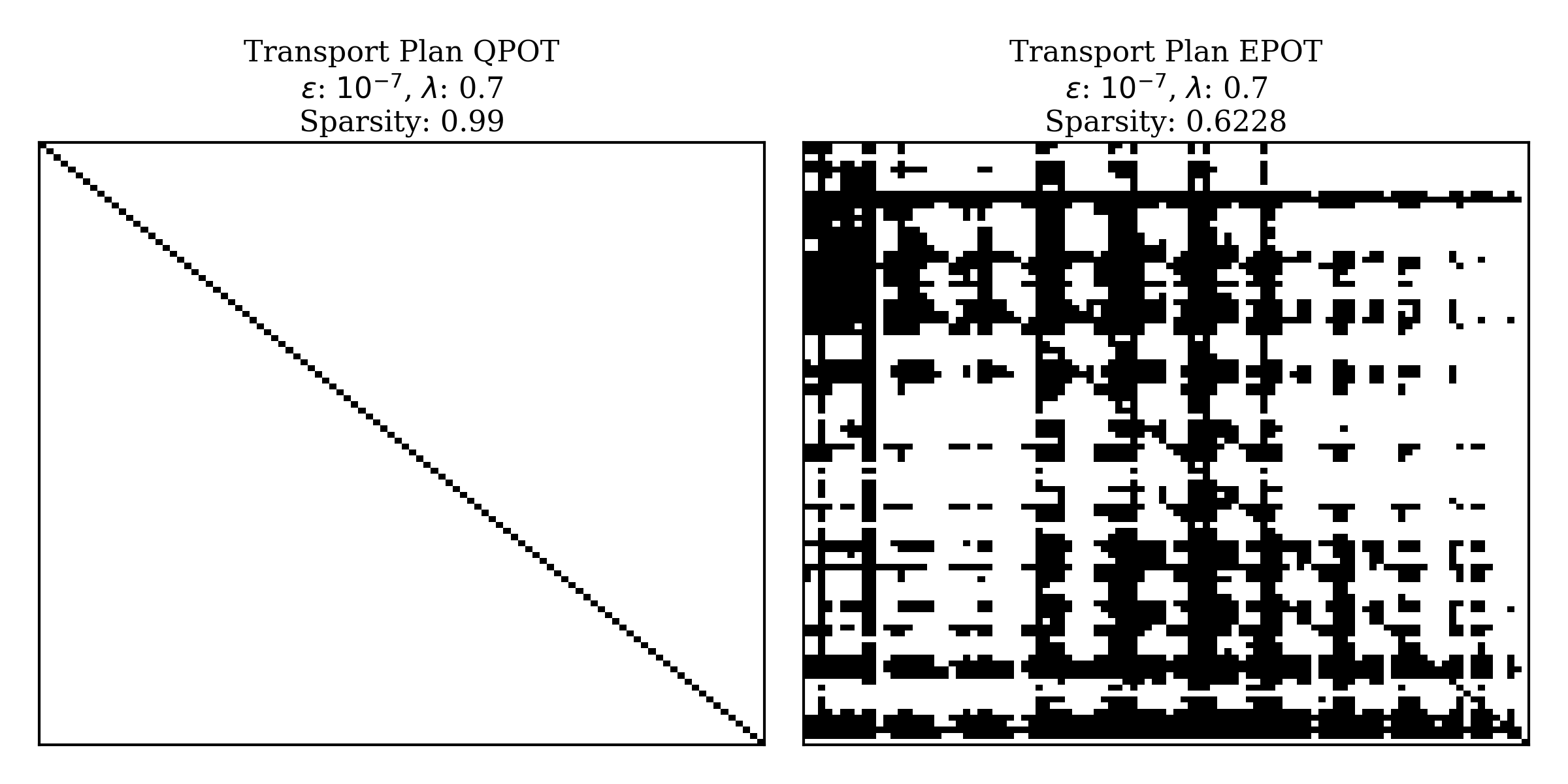}
    \caption{Transport map for QPOT and EPOT on CIFAR10 with $\lambda=0.7$ and $\varepsilon=10^{-7}$}
    \label{fig:cifar10_sparsity}
\end{figure}

For the experiment, $\lambda=0.7$ is chosen as it yielded the best sparsity in the previous experiment. Moreover, we select the regularizer coefficient $\epsilon \approx 2 \times 10^{-7}$. The heat map of the final result is illustrated in Figure \ref{fig:cifar10_sparsity}.

\par
Just like the aforementioned sections, this experiment has again highlighted the dominance of QPOT in terms of getting a sparse transport plan compared to EPOT. While the EPOT heat map is very dense and barely shows any pattern, the QPOT heat map, on the other hand, has given us a much more recognizable solution.

\subsection{Color Transfer}
\begin{figure}[!ht]
\centering
\includegraphics[width=\linewidth]{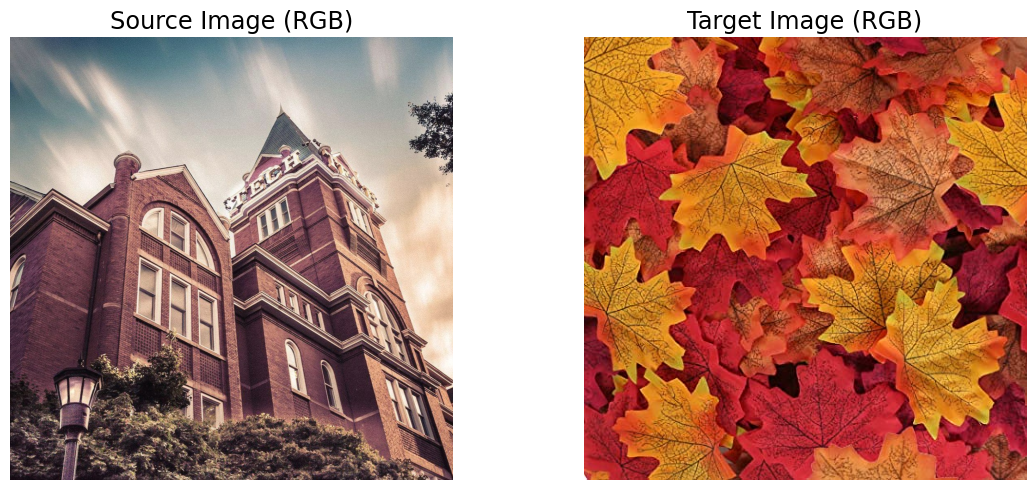}
\caption{Dataset for Color Transfer}
\centering
\label{fig:CFdata_rgb2luv}
\end{figure}

\par
Color transfer is an image editing process where the color palette of one image is applied to another. It is often used to match the mood, tone, or aesthetic of two images by transferring color characteristics (hue, saturation, and brightness) from a reference image to a target image and has applications in robotics and ML \cite{color-transfer-robotics, color-transfer-gamut}. The key to color transfer is that the new image would have the same color palette as the target image without losing its contextual information.
\par
Define an RGB image as $\mathbf{x} \in \mathbb{R}^{h \times w \times 3}$ with each color channel is a matrix of $\mathbb{R}^{h \times w}$ corresponding to Red, Green, or Blue. A pixel $x_{ij}$ in an image at coordinate $(i, j)$ hence consists of three values $x_{ij} = \{ \text{red, green, blue}\}$ where each value range from 0 to 255, where the higher the value, the stronger the color component for that channel. Hence, each image can be interpreted as an array of pixels, with each color channel forming a color distribution. Therefore, the goal of the color transfer task is equivalent to transporting the color distribution of the source image to the target image.

\par 
The data for this experiment is a pair of images (Figure \ref{fig:CFdata_rgb2luv}) representing the source and the target distribution, each having the dimension of $256 \times 256$. This pair of images is chosen because of their distinct color palettes. However, instead of conducting the experiments on the RGB images, we convert the images to the LUV color space for two main reasons. Firstly, the LUV color space primarily encodes the color information as the U and V channels, consequently reducing the dimensionality of the color space from three (Red-Green-Blue) to two (UV-color chromaticity). Moreover, LUV is more accurate in representing the colors, which is in contrast to RGB, where two colors can be close in the color space but appear very different to the human eye. The conversion of RGB to LUV is as follows: $L = R+G+B$, $U=G/L$, and $V=B/L$. 

After converting the color space, we extract the color histogram with \(n\) bins for each image by taking the frequency of each U and V channel. These color histograms then undergo smoothing and normalization, yielding the total mass of each histogram to $\|\mathbf{r}\|_1 = 1$ and $\|\mathbf{c}\|_1 = 1$. From the histogram, we then defined the cost matrix $\mathbf{C}$ as $C_{i, j} = \|a_i - b_j\|^2_2$ where $a_i \in \{a_1, ..., a_n\}$ and $b_j \in \{b_1, ..., b_n\}$ are the bin values of each histogram. Since we are using $256 \times 256$ pictures, setting $\varepsilon$ too small would lead to numerical problems while calculating. Thus, we have chosen to set $i,j$ to $-0.2$ and $-6$ respectively, and use the step size of $10^{-0.2}$.
\par
\begin{figure}[!ht]
    \centering
    \subfloat[Sparsity graph $\lambda=0.7$]{\includegraphics[width=\linewidth]{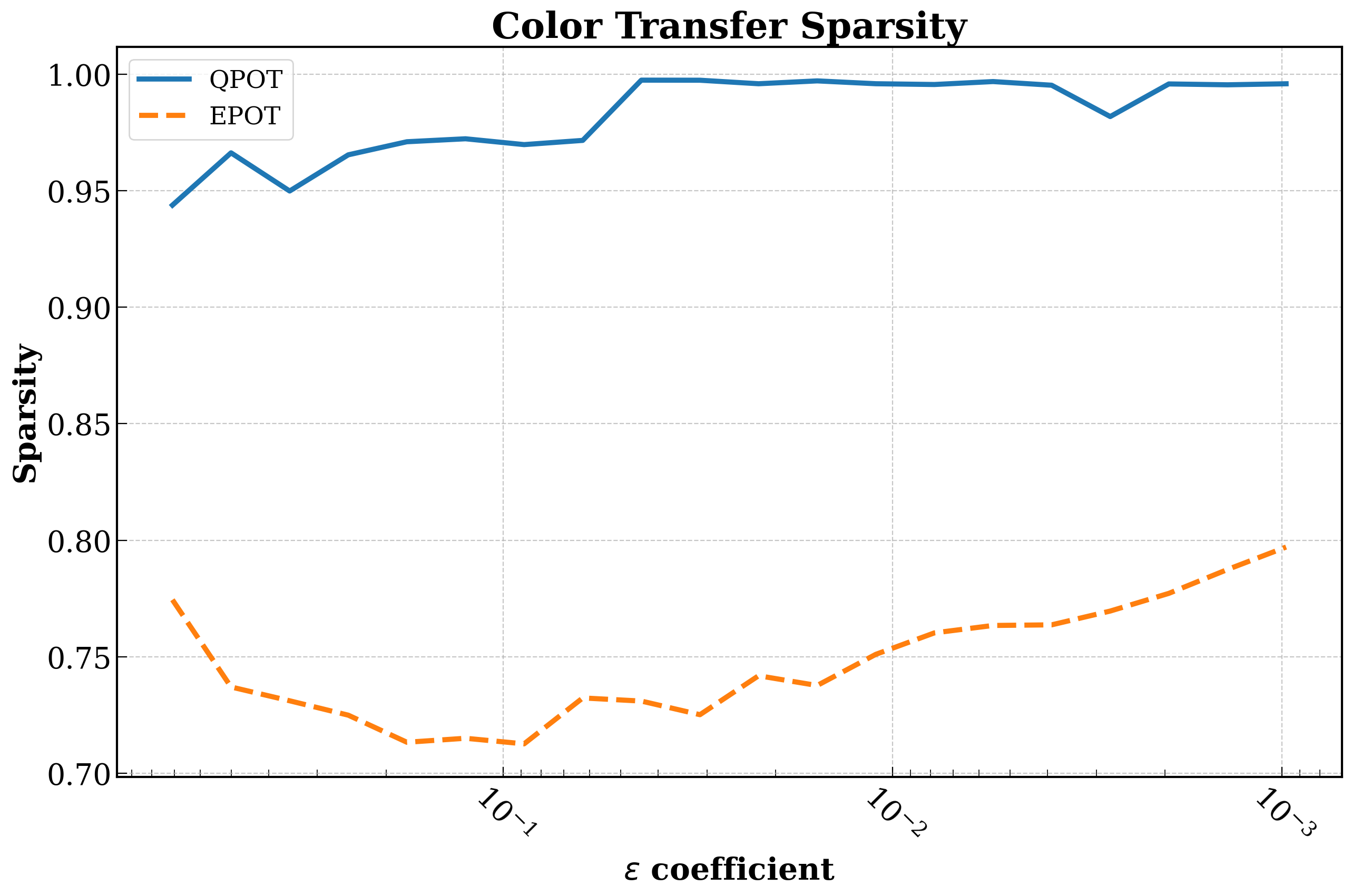}}
    \vspace{1em}
    \subfloat[Transport plans of QPOT and EPOT $\lambda=0.7$]{    \includegraphics[width=\linewidth]{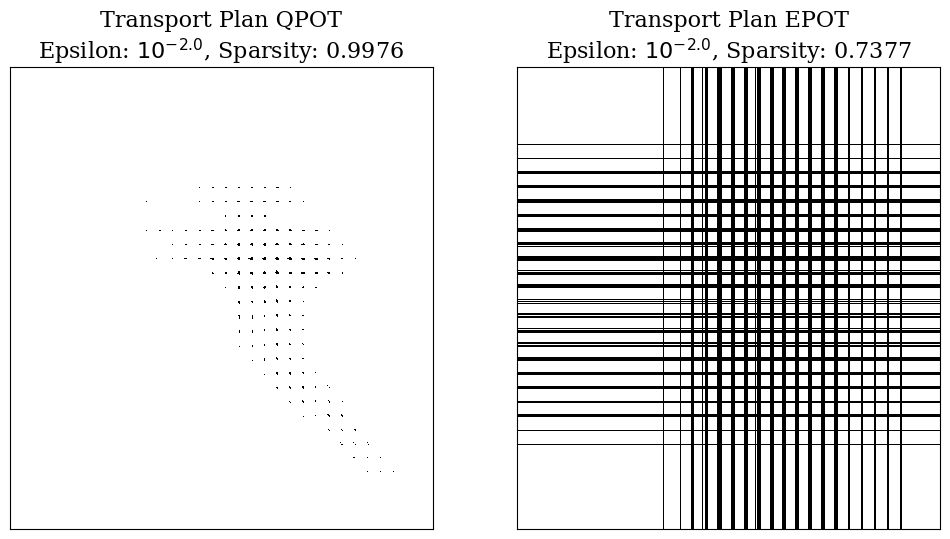}}
    \caption{(a) Sparsity, and (b) Transport plan comparison of QPOT and EPOT for Color Transfer with $\lambda=0.7$}
    \label{fig:EPOT vs QPOT CF}
\end{figure}

The resulting sparsity is seemingly the same for all experimented $\lambda$. We have chosen to plot out the case of $\lambda=0.7$ in Figure \ref{fig:EPOT vs QPOT CF}. It can be observed that QPOT completely dominates EPOT in terms of sparsity, ranging over all the values of $\varepsilon$. This greatly highlights the effectiveness of QPOT in getting a sparse transportation plan. In Figure \ref{fig:EPOT vs QPOT CF} we have also drawn the heatmaps of the results for $\varepsilon=10^{-2}$ of the two methods. These heat maps are drawn by first setting all the values of the transported plan from 0 to 1e-10 to 0 and the rest to 1. The zero-values, then, are plotted white, and the one-values are plotted black. It can be seen that QPOT has a much more sparse transport plan, which clearly shows the pattern of the solution. The heat maps of EPOT, on the other hand, are very dense and have no pattern whatsoever. This firmly confirms the effectiveness of QPOT in terms of sparsity.

\subsection{Domain Adaptation}

\par
In many real-world applications, the process of labeling a newly collected and unlabeled data set (or target domain) requires the usage of models trained on largely annotated data sets (or source
domains). However, problems such as the differences in the probability distribution can impede the effectiveness of directly applying the learned models to the latter. In recent years, new ML research has been conducted to resolve this problem. As a result, Domain Adaptation (DA), a family of techniques that handles cases where source and target samples follow different probabilities, was invented to adapt models trained from the source data onto the target data.
\par
 
\begin{figure}[!ht]
    \centering
    \includegraphics[width=\linewidth]{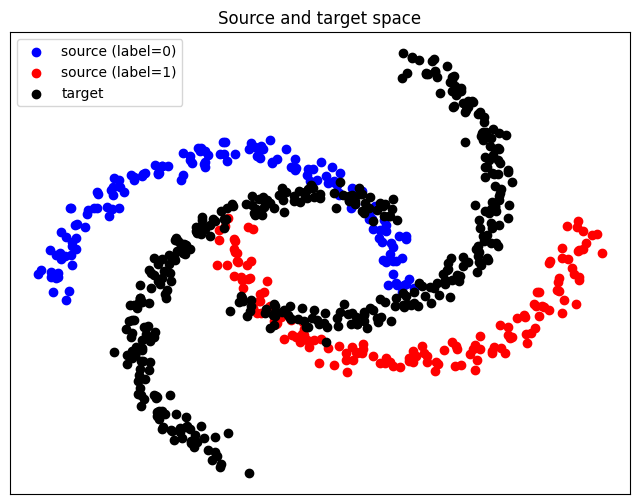}
    \caption{Moon Dataset for domain adaptation}
    \label{fig:DA dataset}
\end{figure}

\par
In our experiment's setting, the "moon" dataset 
, a widely-used scikit-learn \citep{pedregosa2011scikit} synthetic dataset for benchmarking classification algorithms on linearly non-separable data, is used for both the source dataset and the target dataset. This dataset consists of two interleaving half circles representing two distinct classes, commonly used to test algorithms capable of handling complex, non-linear classification boundaries. The experiment requires a source and a target dataset where the target distribution is created by rotating the source distribution by 50 degrees, simulating a domain covariate shift.

\par
Figure \ref{fig:DA dataset} illustrates the dataset where the source dataset has a total of 300 data points, with 150 data points for each class. The target dataset, on the other hand, has a total of 400 data points that are yet to be labeled. Since the number of data points in the target dataset is larger than the amount in the source datasets, we first use K-means clustering to reduce the number of data points in the target dataset down to $300$. The cost matrix is then calculated based on the Euclidean distance and is normalized by dividing all numbers by the largest number. 

\par
The metric for measuring QPOT and EPOT performance is typically the sparsity. However, in this experiment, to measure how accurately our domain adaptation performed, we train a support vector machine (SVM) model on the target dataset predicted by our domain adaptation and then test it on the source dataset to find the accuracy. This accuracy shows how similar the classification of the target dataset is compared to the classification of the source dataset.

\par
\begin{figure}[!ht]
    \centering
    \subfloat[Sparsity $\lambda = 0.7$ and $\varepsilon=10^{-4}$]{\includegraphics[width=\linewidth]{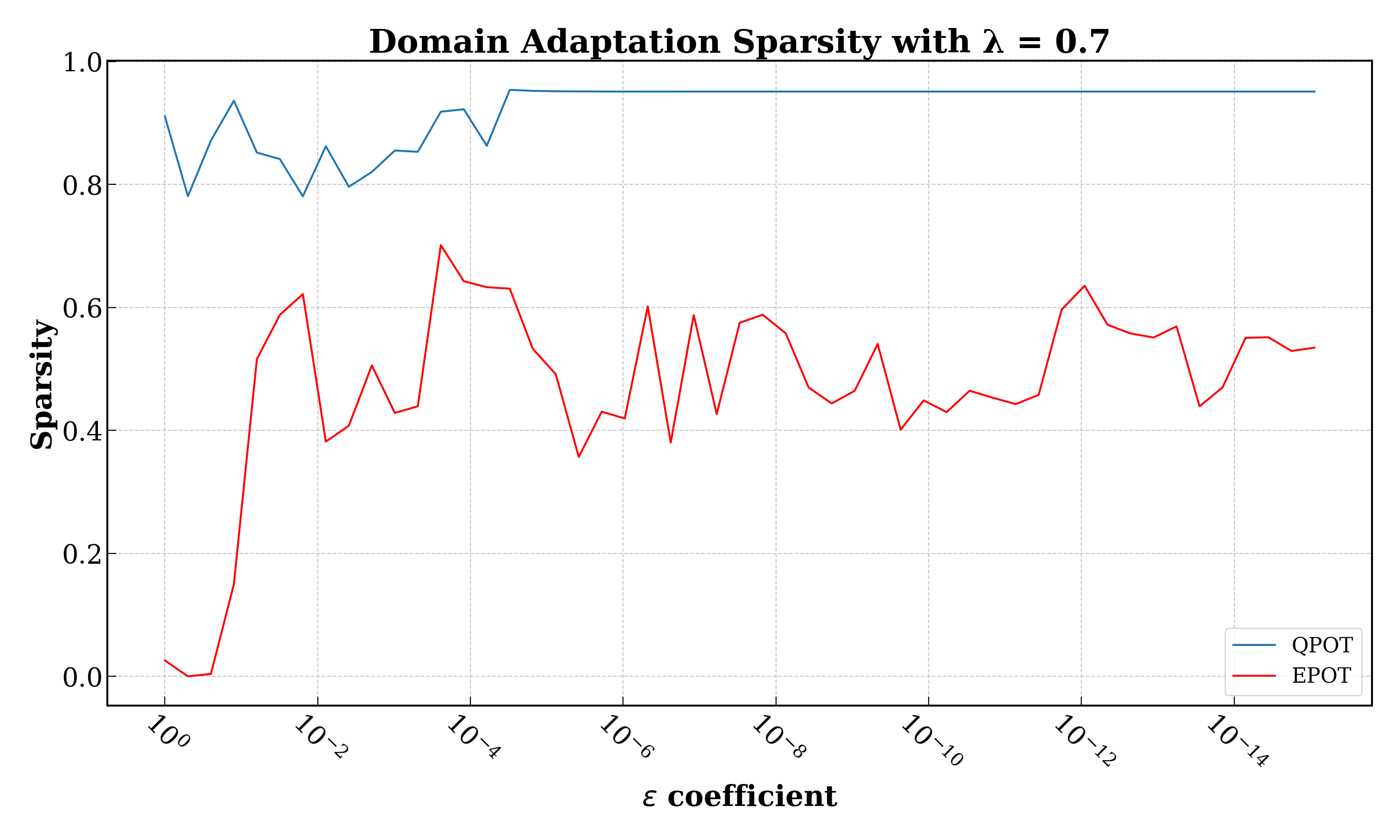}}
    \vspace{1em}
    \centering
    \subfloat[Accuracy $\lambda = 0.7$ ]{\includegraphics[width=\linewidth]{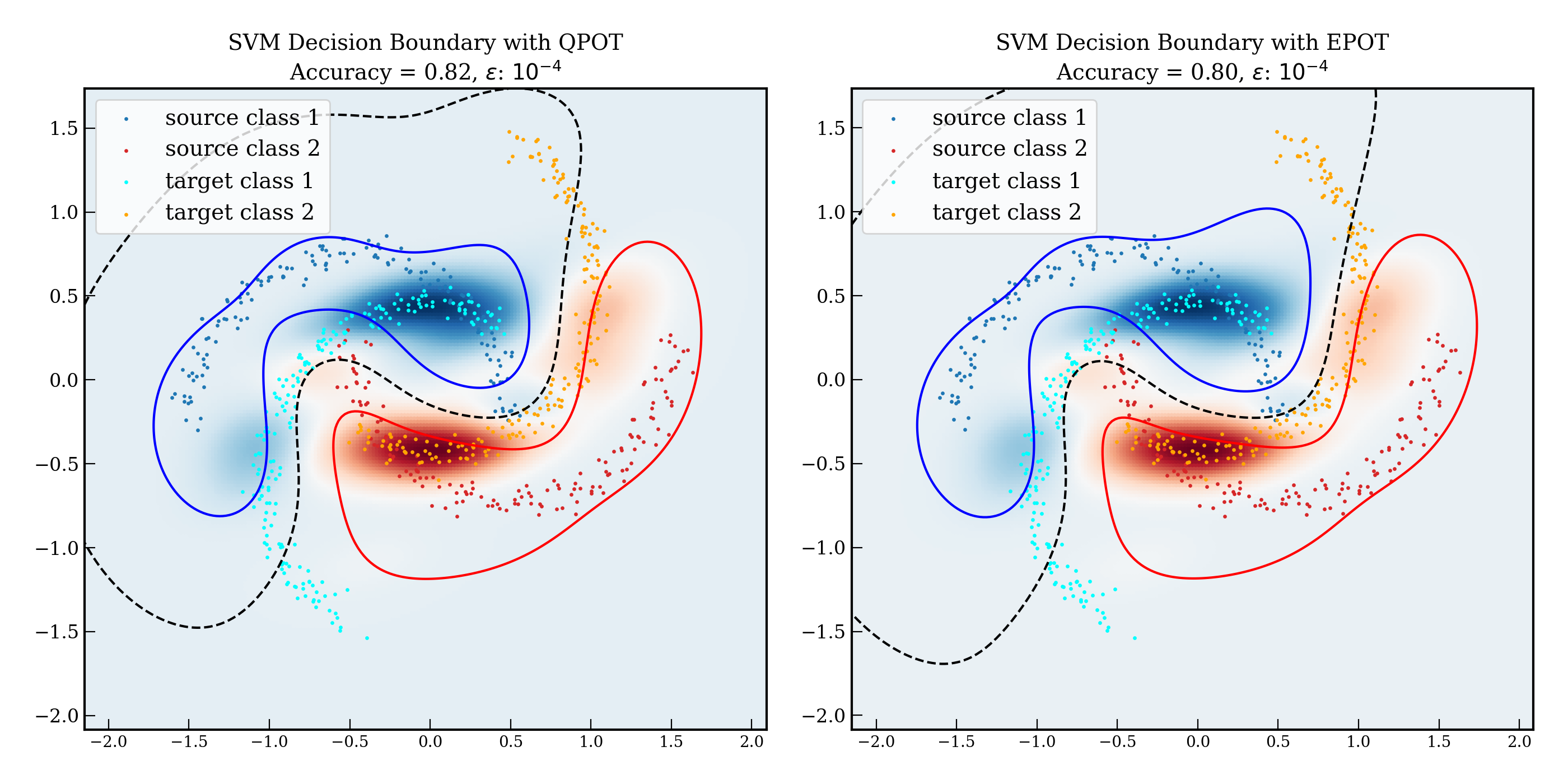}}
    \vspace{1em}
    \centering
    \subfloat[Transport Plans QPOT and EPOT $\lambda = 0.7$ and $\varepsilon=10^{-4}$]{\includegraphics[width=\linewidth]{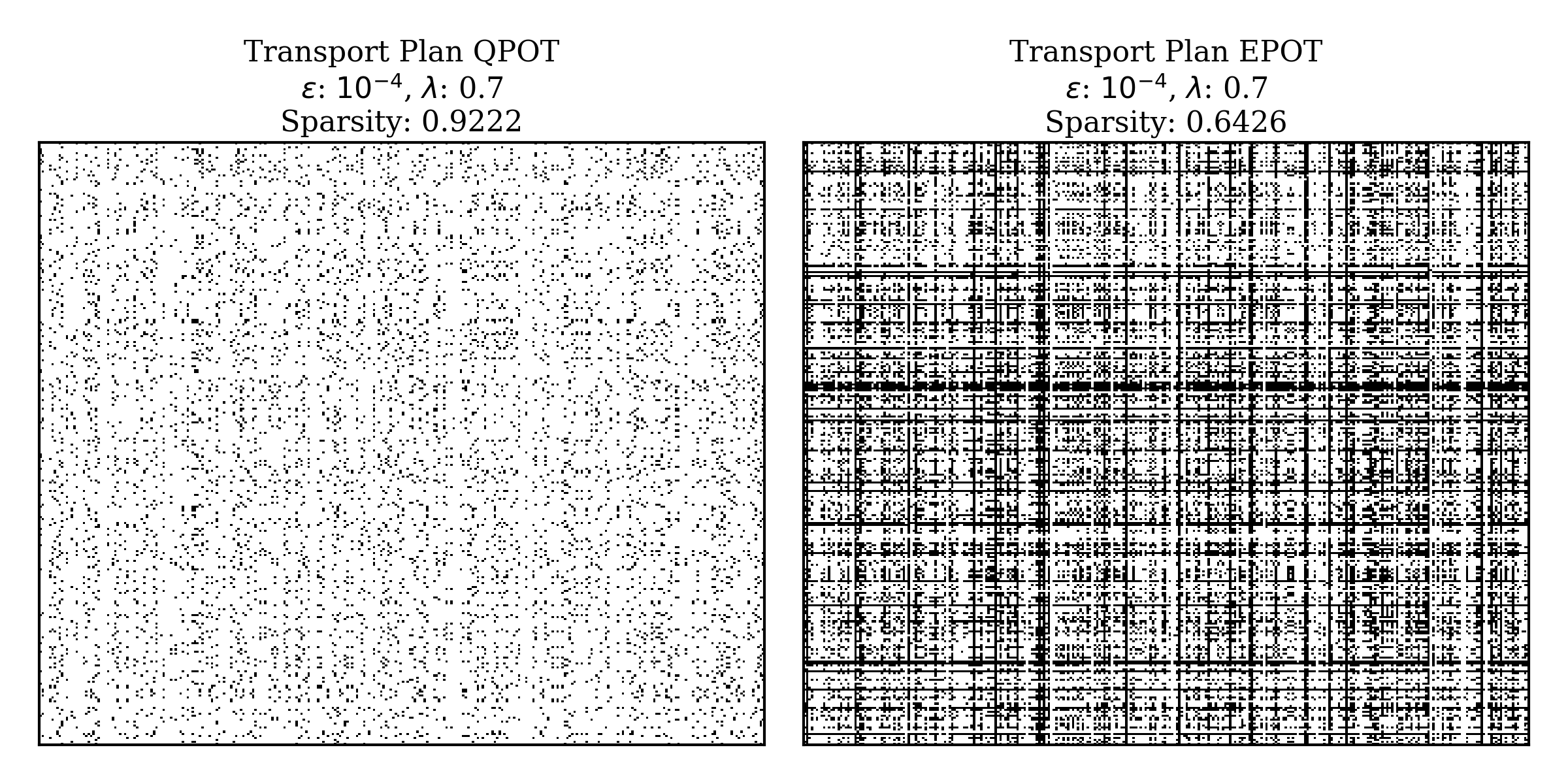}}
    \caption{(a) Sparsity (b) Accuracy with decision boundary of SVM, and (c) Transport plan comparison of QPOT and EPOT for Domain Adaptation with $\lambda = 0.7$.}
    \label{fig:domain_adaptation_result}
\end{figure}

The results of benchmarking QPOT and EPOT on domain adaptation with $\varepsilon$ varying from $10^{-0.3}$ to $10^{-15}$ and $\lambda = 0.7$ are illustrated in Figure \ref{fig:domain_adaptation_result}. The Figure suggests that the sparsity QPOT achieves, averaging around $0.9$, dramatically surpasses that of EPOT, which typically hovers around $0.5$ and peaks at around $0.7$. This performance of sparsity applies to all $\lambda$ values in the experiment. The accuracy of the SVM classifier also shows that there is no tradeoff between sparsity and accuracy from QPOT, where the method provides a good accuracy consistently over the $\varepsilon$ range. Specifically, the accuracies of both solvers are illustrated in Figure \ref{fig:domain_adaptation_result} in the case of $\varepsilon=10^{-4}$, showing an improvement in accuracy from QPOT over EPOT while maintaining high sparsity. Moreover, the heat maps of both methods have also been plotted in Figure \ref{fig:domain_adaptation_result} to further highlight the dominant sparsity of the transport plan created by QPOT compared to that created by EPOT.
\section{Discussion}
As shown in Figure \ref{fig: toy_distribution_sparsity}, Figure \ref{fig:cifar10_sparsity}, Figure \ref{fig:EPOT vs QPOT CF}, and Figure \ref{fig:domain_adaptation_result}(a, c), introducing quadratic terms into the formulation of Partial Optimal Transport (POT) significantly improves sparsity compared to other variants for various tasks. Moreover, QPOT demonstrates strong robustness across different distribution pairs (Figure \ref{fig:BestDisCompare}) and varying transport mass levels (Figure \ref{fig:Lambda best}). Notably, in real-world tasks such as color transfer and domain adaptation, QPOT achieves over a 20\% improvement in performance compared to its counterpart, EPOT. In particular, for domain adaptation, where accuracy serves as an important metric, QPOT also outperforms EPOT (Figure \ref{fig:domain_adaptation_result}(b)), further demonstrating its effectiveness. These results underscore the potential of our approach as a strong foundation for future advancements in sparse OT frameworks. We believe this formulation opens up new directions for both theoretical exploration and practical applications, especially in areas requiring sparse and scalable solutions.

\section{Conclusions and Future Works}
\par In this paper, we proposed and examined the Quadratic-regularized Partial Optimal Transport formulation (QPOT) and implemented a variety of experiments to show its superiority compared to the classical Entropic-regularized Partial Optimal Transport (EPOT) formulation. In particular, we experimented with both synthetic data, the CIFAR-10 dataset, and then applied our formulation to the Domain Adaptation and Color Transfer applications.  For each experiment, we tested different values of the transport mass and the regularizer coefficient to show the proficiency and robustness of QPOT.

\par While entropic regularization is widely favored for its compatibility with the celebrated Sinkhorn algorithm, recent advancements in gradient-based methods suggest that quadratic regularization offers a more seamless integration. Notably, works such as \citep{ blondel-smooth-and-sparse, nguyen2024unbalancedoptimaltransportgradient} have demonstrated the superior efficiency of quadratic regularization in practice. Consistent with these findings, our experiments show that QPOT consistently outperforms EPOT across all settings in terms of sparsity, producing transport plans that make it significantly easier to infer general patterns.

We believe that the new QPOT formulation will lay the foundation for many interesting future works. For example, the sparsity and numerical stability of the QPOT could further enhance the performance and thus facilitate its adoption in many ML and generative AI applications \citep{timeseries1, timeseries2, gilot-interpret-llm}. 
Another approach for future improvements would be to develop accelerated computational methods, such as APDAGD \citep{dvurechensky2018computational}, to solve QPOT. Additionally, one can adapt Stochastic or Constrained Decentralized Optimization methods \citep{wai-defw, hoang-i-pds, nguyen2018sgdhogwildconvergencebounded, zhang2022efficienthpralgorithmwasserstein} to develop sample-efficient computational methods for noisy, dynamic and multi-agent applications \citep{9140785, codedQR, nguyen2023codedcomputingfaulttolerantparallel, Nguyen-Vinh2024}  arising in modern distributed settings  \citep{Nguyen2021UnderstandingTS, 10097233}.



\section{Acknowledgement}
We thank the Journal of Computer Science for providing the resources and platform that made this publication possible. The journal offered us not only a place to share our findings, but also a space where our interest in the topic had room to grow and be expressed. We appreciate the editorial team’s support throughout the review process.

We also extend our sincere thanks to VietDynamic and Binh Duong University for their valuable technical insight and support in shaping the problem formulation. Their contributions were instrumental in guiding the experiments toward meaningful real-world applications and in scaling the method to larger settings.

\section{Authors' Contribution}
The project was a collaborative effort involving contributors with diverse academic backgrounds. Khang Tran and Khoa Nguyen equally co-wrote the manuscript and carried out experiments on synthetic data, Color Transfer, and Domain Adaptation applications. Anh Nguyen and Mai Tran focused on the CIFAR-10 experiments, while Thong Huynh executed and supported the Domain Adaptation experiments. Son Pham, Manh Pham and Mai Ngoc Tran contributed to the QPOT formulation, manuscript writing and future directions. Sy-Hoang Nguyen-Dang assisted with Domain Adaptation experiments. Bang Vo optimized the experiments for large-scale settings. Dung Luong helped run the Color Transfer experiments on large-scale settings and advised on technical matters and industrial applications.

\section{Funding Information}
We want to express our gratitude to VietDynamic and Binh Duong University for their support of this research.

\section{Ethics}
This manuscript is the authors' original work and has not been previously published elsewhere.

\phantomsection
\setlength{\bibsep}{0.5pt} 
\bibliographystyle{apalike} 
\bibliography{references}

\end{document}